\documentclass{article}

\usepackage{arxiv}

\usepackage[utf8]{inputenc} % allow utf-8 input
\usepackage[T1]{fontenc}    % use 8-bit T1 fonts
\usepackage{hyperref}       % hyperlinks
\usepackage{url}            % simple URL typesetting
\usepackage{booktabs}       % professional-quality tables
\usepackage{amsfonts}       % blackboard math symbols
\usepackage{nicefrac}       % compact symbols for 1/2, etc.
\usepackage{microtype}      % microtypography
\usepackage{lipsum}		% Can be removed after putting your text content
\usepackage{graphicx}
\usepackage[numbers]{natbib}
\usepackage{doi}

\title{GenoML: Automated Machine Learning for Genomics}

\usepackage[auth-lg]{authblk}
\usepackage{xpatch}
\xpatchcmd{\author}{\relax#1\relax}{\relax\detokenize{#1}\relax}{}{}
\date{} 					% Or removing it

\author[1,2,3]{Mary B. Makarious}
\author[1,4,5,6]{Hampton L. Leonard}
\author[5]{Dan Vitale}
\author[1,4,5]{Hirotaka Iwaki}
\author[7]{David Saffo}
\author[1,4,8,9]{Lana Sargent}
\author[10]{Anant Dadu}
\author[11]{Eduardo Salmer\'{o}n Casta\~{n}o}
\author[12]{John F. Carter}
\author[13]{Melina Maleknia}
\author[11,14]{Juan A. Botia}
\author[1]{Cornelis Blauwendraat}
\author[10]{Roy H. Campbell}
\author[10]{Sayed Hadi Hashemi}
\author[1,4]{Andrew B. Singleton}
\author[\empty]{%
Mike A. Nalls{\normalfont\textsuperscript{1,4,5}}\thanks{These authors contributed equally to the work.} }
\author[\empty]{%
Faraz Faghri{\normalfont\textsuperscript{1,4,5}}\footnote[1] }

\affil[1]{Laboratory of Neurogenetics, National Institute on Aging, National Institutes of Health, Bethesda, MD, USA}
\affil[2]{Department of Clinical and Movement Neurosciences, UCL Queen Square Institute of Neurology, London, UK}
\affil[3]{UCL Movement Disorders Centre, University College London, London, UK}
\affil[4]{Center for Alzheimer’s and Related Dementias, National Institutes of Health, Bethesda, MD, USA}
\affil[5]{Data Tecnica International LLC, Glen Echo, MD, USA}
\affil[6]{German Center for Neurodegenerative Diseases (DZNE), T\"{u}bingen, Germany}
\affil[7]{Khoury College of Computer Sciences, Northeastern University, Boston, MA, USA}
\affil[8]{School of Nursing, Virginia Commonwealth University, Richmond, VA, USA}
\affil[9]{Geriatric Pharmacotherapy Program, School of Pharmacy, Virginia Commonwealth University, VA, USA}
\affil[10]{Department of Computer Science, University of Illinois at Urbana-Champaign, Urbana, IL, USA
}
\affil[11]{Departamento de Ingenier\'{i}a de la Informaci\'{o}n y las Comunicaciones, Universidad de Murcia, Spain}
\affil[12]{ModelOp, Chicago, IL, USA}
\affil[13]{Georgia Institute of Technology, Atlanta, GA, USA}
\affil[14]{Department of Molecular Neuroscience, UCL Queen Square Institute of Neurology, London, UK}

\date{}

\renewcommand{\headeright}{}
\renewcommand{\undertitle}{}
\hypersetup{
pdftitle={GenoML: Automated Machine Learning for Genomics},
pdfkeywords={Genomics, multi-omics, machine learning, AutoML},
}

\begin{document}
\maketitle

\begin{abstract}
\textit{GenoML} is a Python package automating machine learning workflows for genomics (genetics and multi-omics) with an open science philosophy. Genomics data require significant domain expertise to clean, pre-process, harmonize and perform quality control of the data. Furthermore, tuning, validation, and interpretation involve taking into account the biology and possibly the limitations of the underlying data collection, protocols, and technology. GenoML’s mission is to bring machine learning for genomics and clinical data to non-experts by developing an easy-to-use tool that automates the full development, evaluation, and deployment process. Emphasis is put on open science to make workflows easily accessible, replicable, and transferable within the scientific community. Source code and documentation is available at  \url{https://genoml.com}.
\end{abstract}

% keywords can be removed

\section{Introduction}
In recent years, the demand for machine learning (ML) expertise has outpaced the supply, despite the surge of people entering the field. To address this gap, there have been significant strides in the development of user-friendly machine learning software that can be used by non-experts. The first steps toward simplifying machine learning involved developing simple, unified interfaces to a variety of machine learning algorithms (e.g., scikit-learn \cite{Pedregosa2011nx}, XGBoost \cite{Chen2016-kq}, LightGBM \cite{Ke2017-hj}, TensorFlow \cite{abadi2016tensorflow}, PyTorch \cite{paszke2019pytorch}). Although these packages have made it easy to experiment with machine learning, there is still a fair bit of knowledge and background in data science required to produce high-performing and usable machine learning models. This demand has given rise to the area of automated machine learning (AutoML \cite{Hutter2019-gj}). Some of the recently developed and popular AutoML systems include Auto-WEKA \cite{Thornton2013-sr}, hyperopt-sklearn \cite{komer2014hyperopt}, Auto-sklearn \cite{Feurer2019-xj}, TPOT \cite{olson2016evaluation}, and Auto-Keras \cite{Jin2019-pa}. 

However, different data require different ML pipelines. The development of ML models for genomics (genetics and multi-omics) data, in particular, is notoriously difficult for a non-expert. These data modalities require significant domain expertise to clean, pre-process, harmonize and perform quality control (QC) \cite{Eraslan2019-bf}. Furthermore, tuning, validation, and interpretation involve taking into account the biology and the limitations of the underlying data collection, protocols, and technology.

For ML to truly be accessible to non-experts in the genomics and clinical research areas, we have designed an easy-to-use tool, called GenoML, that automates the full development, evaluation, and deployment process. GenoML provides an end-to-end framework for genomic datasets, including the most complex parts of the process, such as data pre-processing and cleaning, to more advanced training and tuning. GenoML intelligently explores many possible techniques to find the best model for the specific input data. GenoML is also helpful to advanced users; it provides a high-level wrapper performing many modeling tasks that would typically require many more lines of code.

Furthermore, GenoML is more than a package. Since its inception, it has evolved into a diverse community with integrative expertise in data science, bioinformatics, computer science, software engineering, biology, and healthcare. GenoML contributors are staunch advocates of open science, striving to make data and code easily accessible to the scientific community. Please join us and contribute to the development of GenoML.

\section{GenoML Principles and Philosophy}
GenoML developers advocate open science. The following are the underlying principles of GenoML development: 
\begin{itemize}
    \item Little to learn - The goal of GenoML is to democratize complex genomics and machine learning workflows. Thoughtfully designed for newcomers, if a user can `\texttt{cd}' or `\texttt{ls}', they should be able to use GenoML.
    \item Intuitive - Everything has to be simple, straightforward, and effective, from data munging to a tuned model in a few lines of code.
    \item Layered architecture - GenoML is more than a tool; it is an ecosystem that will continuously grow, experimenting with new ideas and innovations. Workflows are kept in logical layers; to change or update one module and not affect the others.
    \item Intelligent defaults - Systematic research is done to set optimized defaults for varying inputs. The default settings are sensible and validated for most workflows to keep modules un-cluttered and to run smoothly. At the same time, providing manual options for advanced users.
    \item No vendor lock-in - Integration with other code, products, and platforms should be hassle-free. GenoML is open source and will remain free and public under the Apache 2.0 license.
    \item Safe and inclusive community - GenoML is a community for positivity in research and open science for the public good. Code of Conduct is adapted from the Contributor Covenant, version 2.0 \cite{noauthor_undated-qi}.  
\end{itemize}

\section{Project Vision}
We foresee a GenoML expansion from primarily an AutoML to a more broadly applicable framework. The next phases of GenoML focus on building an ecosystem of machine learning tools for genomics. Other components would include: 
\begin{itemize}
    \item[--] \textit{GenoML Genetics}: general genetics pipeline tools 
    \item[--] \textit{GenoML Deploy}: designed for deploying the ML models for inference in practice
    \item[--] \textit{GenoML Portal}: an interface designed for clinicians/physicians for use in practice. It also provides model explanation information
    \item[--] \textit{GenoML Federated}: federated learning of GenoML. A critical component in light of recent privacy regulations such as GDPR \cite{Parliament2016-pc}. Enables learning across multiple data silos
    \item[--] \textit{GenoML Meta}: meta-learning aspect of GenoML. Enabling learning and selection across data across diverse datasets and study populations
    \item[--] \textit{GenoML Python}: library developed to integrate seamlessly with other scientific Python libraries
    \item[--] \textit{GenoML Higher API}: higher-level APIs, enabling more community developments
    \item[--] \textit{GenoML Model Zoo}: a place to share models trained on public or private datasets
\end{itemize}

\section{Conclusion}
GenoML automates a wide variety of machine learning pipelines for genomics and multi-omics. Since it relies on other open-source Python packages, it can easily be integrated into existing systems and analytical protocols. We hope for broad adoption and contributions by the community, enabling more unified pipelines producing transparent and reproducible results. Future work includes expansion from primarily an AutoML to an ecosystem of machine learning tools for genomics and more general-purpose AutoML in epidemiological and other health-related domains.

\bibliographystyle{unsrtnat}
\bibliography{references}

\begin{thebibliography}{14}
\providecommand{\natexlab}[1]{#1}
\providecommand{\url}[1]{\texttt{#1}}
\expandafter\ifx\csname urlstyle\endcsname\relax
  \providecommand{\doi}[1]{doi: #1}\else
  \providecommand{\doi}{doi: \begingroup \urlstyle{rm}\Url}\fi

\bibitem[Pedregosa et~al.(2011)Pedregosa, Varoquaux, Gramfort, Michel, Thirion,
  Grisel, Blondel, Prettenhofer, Weiss, Dubourg, and {Others}]{Pedregosa2011nx}
Fabian Pedregosa, Ga{\"e}l Varoquaux, Alexandre Gramfort, Vincent Michel,
  Bertrand Thirion, Olivier Grisel, Mathieu Blondel, Peter Prettenhofer, Ron
  Weiss, Vincent Dubourg, and {Others}.
\newblock Scikit-learn: Machine learning in python.
\newblock \emph{the Journal of machine Learning research}, 12:\penalty0
  2825--2830, 2011.

\bibitem[Chen and Guestrin(2016)]{Chen2016-kq}
Tianqi Chen and Carlos Guestrin.
\newblock {XGBoost}: A scalable tree boosting system.
\newblock In \emph{Proceedings of the 22Nd {ACM} {SIGKDD} International
  Conference on Knowledge Discovery and Data Mining}, KDD '16, pages 785--794,
  New York, NY, USA, 2016. ACM.

\bibitem[Ke et~al.(2017)Ke, Meng, Finley, Wang, Chen, Ma, Ye, and
  Liu]{Ke2017-hj}
Guolin Ke, Qi~Meng, Thomas Finley, Taifeng Wang, Wei Chen, Weidong Ma, Qiwei
  Ye, and Tie-Yan Liu.
\newblock {LightGBM}: A highly efficient gradient boosting decision tree.
\newblock In I~Guyon, U~V Luxburg, S~Bengio, H~Wallach, R~Fergus,
  S~Vishwanathan, and R~Garnett, editors, \emph{Advances in Neural Information
  Processing Systems 30}, pages 3146--3154. Curran Associates, Inc., 2017.

\bibitem[Abadi et~al.(2016)Abadi, Barham, Chen, Chen, Davis, Dean, Devin,
  Ghemawat, Irving, Isard, et~al.]{abadi2016tensorflow}
Mart{\'\i}n Abadi, Paul Barham, Jianmin Chen, Zhifeng Chen, Andy Davis, Jeffrey
  Dean, Matthieu Devin, Sanjay Ghemawat, Geoffrey Irving, Michael Isard, et~al.
\newblock Tensorflow: A system for large-scale machine learning.
\newblock In \emph{12th $\{$USENIX$\}$ symposium on operating systems design
  and implementation ($\{$OSDI$\}$ 16)}, pages 265--283, 2016.

\bibitem[Paszke et~al.(2019)Paszke, Gross, Massa, Lerer, Bradbury, Chanan,
  Killeen, Lin, Gimelshein, Antiga, et~al.]{paszke2019pytorch}
Adam Paszke, Sam Gross, Francisco Massa, Adam Lerer, James Bradbury, Gregory
  Chanan, Trevor Killeen, Zeming Lin, Natalia Gimelshein, Luca Antiga, et~al.
\newblock Pytorch: An imperative style, high-performance deep learning library.
\newblock \emph{Advances in Neural Information Processing Systems},
  32:\penalty0 8026--8037, 2019.

\bibitem[Hutter et~al.(2019)Hutter, Kotthoff, and Vanschoren]{Hutter2019-gj}
Frank Hutter, Lars Kotthoff, and Joaquin Vanschoren.
\newblock \emph{Automated Machine Learning: Methods, Systems, Challenges}.
\newblock Springer, May 2019.

\bibitem[Thornton et~al.(2013)Thornton, Hutter, Hoos, and
  Leyton-Brown]{Thornton2013-sr}
Chris Thornton, Frank Hutter, Holger~H Hoos, and Kevin Leyton-Brown.
\newblock {Auto-WEKA}: combined selection and hyperparameter optimization of
  classification algorithms.
\newblock In \emph{Proceedings of the 19th {ACM} {SIGKDD} international
  conference on Knowledge discovery and data mining}, KDD '13, pages 847--855,
  New York, NY, USA, August 2013. Association for Computing Machinery.

\bibitem[Komer et~al.(2014)Komer, Bergstra, and Eliasmith]{komer2014hyperopt}
Brent Komer, James Bergstra, and Chris Eliasmith.
\newblock Hyperopt-sklearn: automatic hyperparameter configuration for
  scikit-learn.
\newblock In \emph{ICML workshop on AutoML}, volume~9, page~50. Citeseer, 2014.

\bibitem[Feurer et~al.(2019)Feurer, Klein, Eggensperger, Springenberg, Blum,
  and Hutter]{Feurer2019-xj}
Matthias Feurer, Aaron Klein, Katharina Eggensperger, Jost~Tobias Springenberg,
  Manuel Blum, and Frank Hutter.
\newblock Auto-sklearn: efficient and robust automated machine learning.
\newblock In \emph{Automated Machine Learning}, pages 113--134. Springer, Cham,
  2019.

\bibitem[Olson et~al.(2016)Olson, Bartley, Urbanowicz, and
  Moore]{olson2016evaluation}
Randal~S Olson, Nathan Bartley, Ryan~J Urbanowicz, and Jason~H Moore.
\newblock Evaluation of a tree-based pipeline optimization tool for automating
  data science.
\newblock In \emph{Proceedings of the Genetic and Evolutionary Computation
  Conference 2016}, pages 485--492, 2016.

\bibitem[Jin et~al.(2019)Jin, Song, and Hu]{Jin2019-pa}
Haifeng Jin, Qingquan Song, and Xia Hu.
\newblock {Auto-Keras}: An efficient neural architecture search system.
\newblock In \emph{Proceedings of the 25th {ACM} {SIGKDD} International
  Conference on Knowledge Discovery \& Data Mining}, KDD '19, pages 1946--1956,
  New York, NY, USA, July 2019. Association for Computing Machinery.

\bibitem[Eraslan et~al.(2019)Eraslan, Avsec, Gagneur, and
  Theis]{Eraslan2019-bf}
G{\"o}kcen Eraslan, {\v Z}iga Avsec, Julien Gagneur, and Fabian~J Theis.
\newblock Deep learning: new computational modelling techniques for genomics.
\newblock \emph{Nat. Rev. Genet.}, 20\penalty0 (7):\penalty0 389--403, July
  2019.

\bibitem[noa()]{noauthor_undated-qi}
Contributor covenant:.
\newblock
  \url{https://www.contributor-covenant.org/version/2/0/code_of_conduct/}.
\newblock Accessed: 2021-3-3.

\bibitem[Parliament and the European~Union(2016)]{Parliament2016-pc}
European Parliament and The Council~of the European~Union.
\newblock Regulation ({EU}) 2016/679 of the european parliament and of the
  council of 27 april 2016 on the protection of natural persons with regard to
  the processing of personal data and on the free movement of such data, and
  repealing directive {95/46/EC} (general data protection regulation).
\newblock \emph{Official Journal of the European Union OJ}, 59:\penalty0 1--88,
  2016.

\end{thebibliography}

\end{document}